\documentclass[final]{cvpr}

\usepackage{bm}
\usepackage{textcomp} 

\usepackage{multirow}
\usepackage{subcaption}
\usepackage{amssymb} 
\usepackage{pifont} 
\usepackage{makecell} 
\usepackage{amsmath}
\usepackage{amsfonts}
\usepackage{ifthen}

\makeatletter
\@namedef{ver@everyshi.sty}{}
\makeatother
\usepackage{tikz}
\usepackage{tikz/tikzit}

\tikzstyle{pixel coord}=[fill={rgb,255: red,76; green,148; blue,255}, draw=black, shape=circle, scale=0.5]
\tikzstyle{sine}=[fill={rgb,255: red,157; green,255; blue,249}, draw=black, shape=circle]
\tikzstyle{relu}=[fill=white, draw=black, shape=circle]
\tikzstyle{red}=[fill={rgb,255: red,231; green,76; blue,60}, draw=black, shape=circle]
\tikzstyle{green}=[fill={rgb,255: red,46; green,204; blue,113}, draw=black, shape=circle]
\tikzstyle{blue}=[fill={rgb,255: red,52; green,152; blue,219}, draw=black, shape=circle]
\tikzstyle{l1 pixel coord}=[fill={rgb,255: red,114; green,58; blue,255}, draw=black, shape=circle]
\tikzstyle{l2 pixel coord}=[fill={rgb,255: red,139; green,255; blue,153}, draw=black, shape=circle]
\tikzstyle{l1 pixel coord neuron}=[fill={rgb,255: red,114; green,58; blue,255}, draw=black, shape=circle]
\tikzstyle{l2 pixel coord neuron}=[fill={rgb,255: red,139; green,255; blue,153}, draw=black, shape=circle]
\tikzstyle{l3 pixel coord neuron}=[fill={rgb,255: red,76; green,148; blue,255}, draw=black, shape=circle]
\tikzstyle{pixel-circle}=[fill={rgb,255: red,76; green,148; blue,255}, draw=black, shape=circle]
\tikzstyle{gassuain-1}=[fill=none, draw={rgb,255: red,114; green,58; blue,255}, shape=circle, scale=3]
\tikzstyle{gaussian-2}=[fill=none, draw={rgb,255: red,117; green,105; blue,255}, shape=circle, scale=8]
\tikzstyle{gaussian-3}=[fill=none, draw={rgb,255: red,119; green,171; blue,255}, shape=circle, scale=15]

\tikzstyle{thick line}=[-, line width=0.35mm]
\tikzstyle{blue line}=[-, draw={rgb,255: red,51; green,95; blue,255}]
\tikzstyle{blue line dashed}=[-, draw={rgb,255: red,51; green,95; blue,255}, dashed]
\tikzstyle{dashed}=[-, dashed]
\tikzstyle{layer 0 connections}=[-, draw={rgb,255: red,122; green,56; blue,255}, line width=0.3mm]
\tikzstyle{layer 1 connections}=[-, draw={rgb,255: red,117; green,105; blue,255}, line width=0.3mm]
\tikzstyle{layer 2 connections}=[-, draw={rgb,255: red,119; green,171; blue,255}, line width=0.3mm]
\tikzstyle{layer 3 connections}=[-, draw={rgb,255: red,135; green,217; blue,255}, line width=0.3mm]
\tikzstyle{layer 0 params}=[-, fill={rgb,255: red,114; green,58; blue,255}]
\tikzstyle{layer 1 params}=[-, fill={rgb,255: red,117; green,105; blue,255}]
\tikzstyle{layer 2 params}=[-, fill={rgb,255: red,119; green,171; blue,255}]
\tikzstyle{layer 3 params}=[-, fill={rgb,255: red,135; green,217; blue,255}]
\tikzstyle{l2 pixel line}=[-, draw={rgb,255: red,139; green,255; blue,153}]
\tikzstyle{l2 pixel arrow}=[draw={rgb,255: red,139; green,255; blue,153}, ->]
\tikzstyle{l3 pixel arrow}=[draw={rgb,255: red,76; green,148; blue,255}, ->]
\tikzstyle{l1 pixel line}=[-, draw={rgb,255: red,114; green,58; blue,255}]
\tikzstyle{l1 pixel arrow}=[draw={rgb,255: red,114; green,58; blue,255}, ->]
\tikzstyle{arrow}=[->]
\tikzstyle{connection}=[-, draw={rgb,255: red,76; green,148; blue,255}]

\usetikzlibrary{arrows,calc,decorations.markings,math,arrows.meta,decorations.pathreplacing}
\definecolor{mediumtealblue}{rgb}{0.0, 0.33, 0.71}
\definecolor{darkpastelgreen}{rgb}{0.01, 0.75, 0.24}
\definecolor{azure(colorwheel)}{rgb}{0.0, 0.5, 1.0}

\newcommand{\R}{\mathbb{R}}

\newcommand{\expect}[2][]{
\ifthenelse{\equal{#1}{}}{
\mathbb{E}\left[#2\right]
}{
\underset{#1}{\mathbb{E}}\left[#2\right]
}}

\usepackage{times}
\usepackage{epsfig}
\usepackage{graphicx}
\usepackage{amsmath}
\usepackage{amssymb}

\usepackage[pagebackref=true,breaklinks=true,colorlinks,bookmarks=false]{hyperref}

\def\cvprPaperID{8135} 
\def\confYear{CVPR 2021}

\begin{document}

\title{Interpolating Points on a Non-Uniform Grid using a Mixture of Gaussians}

\author{Ivan Skorokhodov\\
KAUST\\
Thuwal, Saudi Arabia\\
{\tt\small iskorokhodov@gmail.com}
}

\maketitle

\begin{abstract}
In this work, we propose an approach to perform non-uniform image interpolation based on a Gaussian Mixture Model.
Traditional image interpolation methods, like nearest neighbor, bilinear, Hamming, Lanczos, etc. assume that the coordinates you want to interpolate from, are positioned on a uniform grid.
However, it is not always the case in practice and we develop an interpolation method that is able to generate an image from arbitrarily positioned pixel values.
We do this by representing each known pixel as a 2D normal distribution and considering each output image pixel as a sample from the mixture of all the known ones.
Apart from the ability to reconstruct an image from arbitrarily positioned set of pixels, this also allows us to differentiate through the interpolation procedure, which might be helpful for downstream applications.
Our optimized CUDA kernel and the source code to reproduce the benchmarks is located at \url{https://github.com/universome/non-uniform-interpolation}.
\end{abstract}

\section{Introduction}


Imagine that we have access to some image in a \textit{functional} form.
I.e. the image is represented as a function $f: \bm{p} \mapsto \bm{c}$ which takes a pixel coordinate $\bm{p} = (x,y) \in \R^2$ as an input and produces its corresponding RGB value $\bm{c} = (r, g, b) \in \R^3$.
Such representations arise, for example, in differentiable rendering pipelines \cite{SynSin, SoftRasterizer} or implicit representations of images \cite{siren, fourier_inr, inr_gan, cips}.

Now imagine, that we want to generate a high-resolution raster image from this functional representation $f(\bm p)$.
This means that we need to evaluate $f(\bm p)$ in every coordinate location of $H \times W$ grid to generate a $H \times W$ sized image.
If evaluating $f(\bm p)$ is costly then it is a tedious procedure.
What can we do?

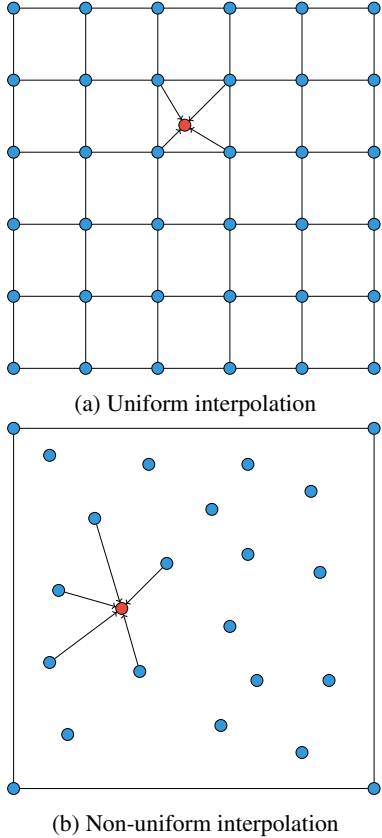
\begin{figure}
\centering
\begin{subfigure}[b]{0.45\textwidth}
\centering
\resizebox{5cm}{!}{\begin{tikzpicture}
	\begin{pgfonlayer}{nodelayer}
		\node [style=blue] (0) at (-5, 5) {};
		\node [style=blue] (1) at (-5, -5) {};
		\node [style=blue] (2) at (5, -5) {};
		\node [style=blue] (3) at (5, 5) {};
		\node [style=blue] (4) at (-3, 5) {};
		\node [style=blue] (5) at (-3, -5) {};
		\node [style=blue] (6) at (-1, 5) {};
		\node [style=blue] (7) at (-1, -5) {};
		\node [style=blue] (8) at (1, 5) {};
		\node [style=blue] (9) at (1, -5) {};
		\node [style=blue] (10) at (3, 5) {};
		\node [style=blue] (11) at (3, -5) {};
		\node [style=blue] (12) at (-5, 3) {};
		\node [style=blue] (13) at (5, 3) {};
		\node [style=blue] (14) at (-3, 3) {};
		\node [style=blue] (15) at (-1, 3) {};
		\node [style=blue] (16) at (1, 3) {};
		\node [style=blue] (17) at (3, 3) {};
		\node [style=blue] (18) at (-5, 1) {};
		\node [style=blue] (19) at (5, 1) {};
		\node [style=blue] (20) at (-3, 1) {};
		\node [style=blue] (21) at (-1, 1) {};
		\node [style=blue] (22) at (1, 1) {};
		\node [style=blue] (23) at (3, 1) {};
		\node [style=blue] (24) at (-5, -1) {};
		\node [style=blue] (25) at (5, -1) {};
		\node [style=blue] (26) at (-3, -1) {};
		\node [style=blue] (27) at (-1, -1) {};
		\node [style=blue] (28) at (1, -1) {};
		\node [style=blue] (29) at (3, -1) {};
		\node [style=blue] (30) at (-5, -3) {};
		\node [style=blue] (31) at (5, -3) {};
		\node [style=blue] (32) at (-3, -3) {};
		\node [style=blue] (33) at (-1, -3) {};
		\node [style=blue] (34) at (1, -3) {};
		\node [style=blue] (35) at (3, -3) {};
		\node [style=red] (36) at (-0.25, 1.75) {};
	\end{pgfonlayer}
	\begin{pgfonlayer}{edgelayer}
		\draw (0) to (3);
		\draw (1) to (2);
		\draw (0) to (1);
		\draw (3) to (2);
		\draw (4) to (5);
		\draw (6) to (7);
		\draw (8) to (9);
		\draw (10) to (11);
		\draw (12) to (13);
		\draw (18) to (19);
		\draw (24) to (25);
		\draw (30) to (31);
		\draw [style=arrow] (15) to (36);
		\draw [style=arrow] (21) to (36);
		\draw [style=arrow] (22) to (36);
		\draw [style=arrow] (16) to (36);
	\end{pgfonlayer}
\end{tikzpicture}}
\caption{Uniform interpolation}
\label{fig:uniform-interpolation}
\end{subfigure}
\hfill
\begin{subfigure}[b]{0.45\textwidth}
\centering
\resizebox{5cm}{!}{\begin{tikzpicture}
	\begin{pgfonlayer}{nodelayer}
		\node [style=blue] (0) at (-4.75, 5) {};
		\node [style=blue] (1) at (-4.75, -5) {};
		\node [style=blue] (2) at (5.25, -5) {};
		\node [style=blue] (3) at (5.25, 5) {};
		\node [style=blue] (4) at (-2.5, 2.5) {};
		\node [style=blue] (5) at (-3.5, 0.5) {};
		\node [style=blue] (6) at (-1.25, -1.75) {};
		\node [style=blue] (7) at (-3.25, -3.5) {};
		\node [style=blue] (8) at (-0.5, 1.25) {};
		\node [style=blue] (9) at (-1, 4) {};
		\node [style=blue] (10) at (3.5, 3.25) {};
		\node [style=blue] (11) at (0.75, 2.75) {};
		\node [style=blue] (12) at (3.75, 1) {};
		\node [style=blue] (13) at (1.25, -0.5) {};
		\node [style=blue] (14) at (4, -2) {};
		\node [style=blue] (15) at (1, -3.25) {};
		\node [style=red] (16) at (-1.75, 0) {};
		\node [style=blue] (17) at (-3.75, 4.25) {};
		\node [style=blue] (18) at (-3.75, -1.5) {};
		\node [style=blue] (19) at (2, -2) {};
		\node [style=blue] (20) at (3.25, -4) {};
		\node [style=blue] (21) at (1.75, 1.5) {};
		\node [style=blue] (22) at (1.75, 4) {};
	\end{pgfonlayer}
	\begin{pgfonlayer}{edgelayer}
		\draw (3) to (2);
		\draw (1) to (2);
		\draw (0) to (1);
		\draw (0) to (3);
		\draw [style=arrow] (5) to (16);
		\draw [style=arrow] (8) to (16);
		\draw [style=arrow] (6) to (16);
		\draw [style=arrow] (4) to (16);
		\draw [style=arrow] (18) to (16);
	\end{pgfonlayer}
\end{tikzpicture}}
\caption{Non-uniform interpolation}
\label{fig:non-uniform-interpolation}
\end{subfigure}
\caption{Example of (a) uniform and (b) non-uniform interpolation. Blue points are ``known'' points and red points are ``unknown'' points, i.e. points we want compute the value in. Existing interpolation methods assume a uniform grid, but it is not always true in practice.}
\end{figure}

One approach would be to speed up the inference for $f(\bm p)$.
Another one is to generate a low-resolution version of an image and then upsample it with one of the existing methods.
Such upsampling methods assume that the points you are trying to upsample from, are positioned on a uniform grid, i.e. they have a fixed equal horizontal and vertical spacing between each other, as depicted on figure \ref{fig:uniform-interpolation}.
However, in practice there sometimes occur situations when your points are positioned on a non-uniform grid, like on image \ref{fig:non-uniform-interpolation}, limiting the applicability of the existing tools.

To alleviate the issue, we propose a novel interpolation method that makes it possible to reconstruct an image from a subset of points that are arbitrarily scattered across the image.
We achieve this by representing each known color $c^{i}$ at location $(x^{(i)}, y^{(i)})$ as a 2D normal distribution $\mathcal{N}(\bm \mu^{(i)}, \sigma^2 I)$ for $\bm\mu^{(i)} = (x^{(i)}, y^{(i)})$ and some predefined variance $\sigma^2$.

To summarize, our contributions are the following:
\begin{itemize}
    \item We propose a novel interpolation technique which is based on representing the known points as a GMM model and inferring the value for the unknown ones as an expectation.
    \item We develop an optimized CUDA kernel for both the forward and backward passes of the proposed interpolation procedure.
    \item We conduct the experiments on ImageNet dataset and show that our proposed interpolation technique outperforms in several scenarios 6 other standard interpolation methods based on the reconstruction quality.
\end{itemize}

\section{Method}

Our interpolation method treats each known point $(x^{(i)}, y^{(i)})$ with color $c^{(i)}$ as a 2D gaussian distribution with mean $\bm \mu^{(i)} = (x^{(i)}, y^{(i)})$ and some diagonal covariance matrix $\sigma^2 I$ for some fixed hyperparameter $\sigma$.
To compute the point value in some unknown pixel coordinate position $\bm q = (x, y)$ we evaluate its expected color value as:

\begin{equation}\label{eq:color-value}
c(\bm q) \triangleq \expect[p(c | \bm q)]{c} = \sum_{i=1}^N c(\bm p^{(i)}) \cdot \frac{\mathcal{N}(\bm q | \bm\mu^{(i)}, \sigma^2 I)}{Z_{\bm q}}
\end{equation}
where $Z_{\bm q}$ is the normalizing factor for the point computed as:
\begin{equation}
Z_{\bm q} = \sum_{i=1}^N \mathcal{N}(\bm q | \bm\mu^{(i)}, \sigma^2 I)
\end{equation}

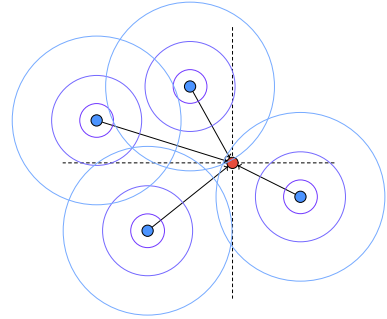
\begin{figure}
\centering
\resizebox{5cm}{!}{\begin{tikzpicture}
	\begin{pgfonlayer}{nodelayer}
		\node [style=pixel-circle] (0) at (-3, 2.25) {};
		\node [style=pixel-circle] (1) at (-1.5, -1) {};
		\node [style=pixel-circle] (2) at (3, 0) {};
		\node [style=red] (3) at (1, 1) {};
		\node [style=pixel-circle] (4) at (-0.25, 3.25) {};
		\node [style=none] (5) at (1, 5) {};
		\node [style=none] (6) at (1, -3) {};
		\node [style=none] (7) at (-4, 1) {};
		\node [style=none] (8) at (4, 1) {};
		\node [style=gassuain-1] (9) at (-0.25, 3.25) {};
		\node [style=gassuain-1] (10) at (-3, 2.25) {};
		\node [style=gassuain-1] (11) at (-1.5, -1) {};
		\node [style=gassuain-1] (12) at (3, 0) {};
		\node [style=gaussian-2] (13) at (-0.25, 3.25) {};
		\node [style=gaussian-2] (14) at (-3, 2.25) {};
		\node [style=gaussian-2] (15) at (-1.5, -1) {};
		\node [style=gaussian-2] (16) at (3, 0) {};
		\node [style=gaussian-3] (17) at (-0.25, 3.25) {};
		\node [style=gaussian-3] (18) at (-3, 2.25) {};
		\node [style=gaussian-3] (19) at (-1.5, -1) {};
		\node [style=gaussian-3] (20) at (3, 0) {};
	\end{pgfonlayer}
	\begin{pgfonlayer}{edgelayer}
		\draw [style=arrow] (4) to (3);
		\draw [style=arrow] (0) to (3);
		\draw [style=arrow] (1) to (3);
		\draw [style=arrow] (2) to (3);
		\draw [densely dashed] (8.center) to (7.center);
		\draw [densely dashed] (5.center) to (6.center);
	\end{pgfonlayer}
\end{tikzpicture}}
\caption{The proposed interpolation method. We treat known (blue) points as gaussian clusters where their mean vectors are specified by their coordinate positions and their variances are fixed. For each unknown (red) pixel we compute its color value as an expectation over all the gaussians using the formula \eqref{eq:color-value}. The closer a point to a given cluster --- the more it influences its resulting color.}
\label{fig:gmm-interpolation}
\end{figure}

To speed up the procedure, we consider only those known points, that are close enough to the query one.
The gaussian densities are considered to be weights by which the known points influence the resulted color of the unknown one.
We illustrate this on Figure~\ref{fig:gmm-interpolation}.

We interpolate each color channel independently.

\section{Experiments}
\subsection{Validating the correctness of the computations}
Since writing CUDA kernels is very error-prone, especially for the backward pass, one needs to ensure that all the computations are correct.
For this, we implemented a (very) slow python version using Pytorch automatic differentiation framework \cite{Pytorch}.
After that, we performed the forward pass on the same input for both the python version and our optimized CUDA kernel.
Comparing that the results of the both procedures are equal, confirms that the implemented computations are correct.

\subsection{Testing the reconstruction quality}
\begin{figure}
\centering
\includegraphics[width=\linewidth]{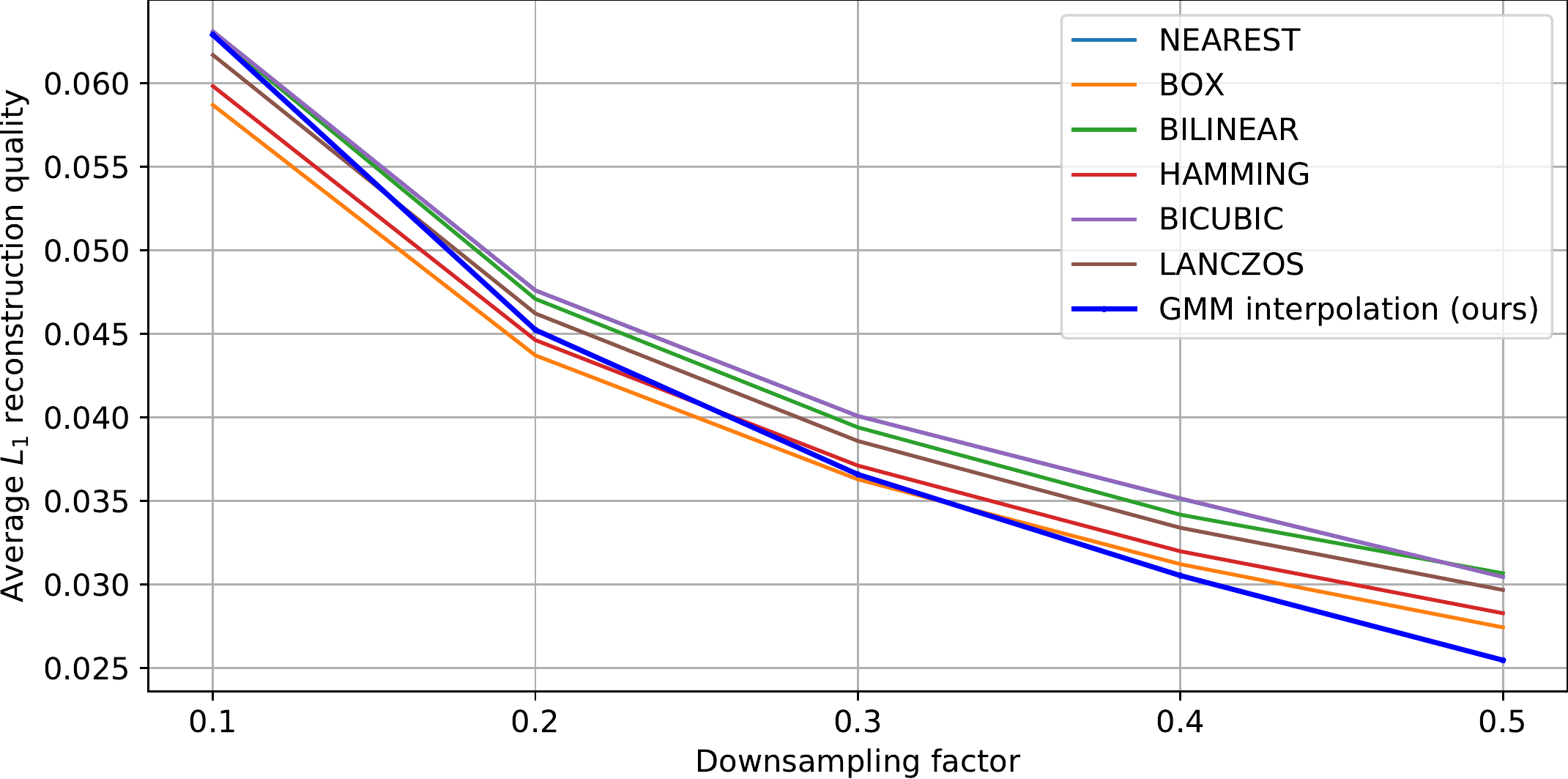}
\caption{Downsampling ImageNet images and then upsampling them with different interpolation methods. Our method starts to outperform existing interpolation techniques in terms of $L_1$ metric when the downsampling factor increases.}
\label{fig:imagenet-experiments}
\end{figure}

The first set of experiments we conduct is to test the reconstruction quality of the proposed interpolation method.
For this, we take 1000 images from ImageNet dataset \cite{ImageNet} --- one image per class --- then downsample them to a specified factor and then upsample with one of the methods.
We test against 6 standard interpolation techniques that are shipped into PIL image library \cite{PIL}: nearest neighbour, box, bilinear, bicubic, Hamming and Lancoz.
The results are presented on Figure~\ref{fig:imagenet-experiments}.
As one can see, our method is competitive for small downsampling factors and outperforms the existing methods when the downsampling factor increases.

\subsection{Optimizing the points locations}
Since our procedure permits the optimization of points positions, it is a natural idea to minimize the reconstruction quality with gradient descent.
Concretely, $\bm \mu^{(1)}, ..., \bm \mu^{(N)}$ become learnable parameters that are being optimized using the derivatives compute with respect to them.

We take an image of a room, define how many points we allow ourselves to have, randomly sample the points on an image using the uniform distribution and then optimize their locations.

\begin{figure}
    \centering
    \includegraphics[width=\linewidth]{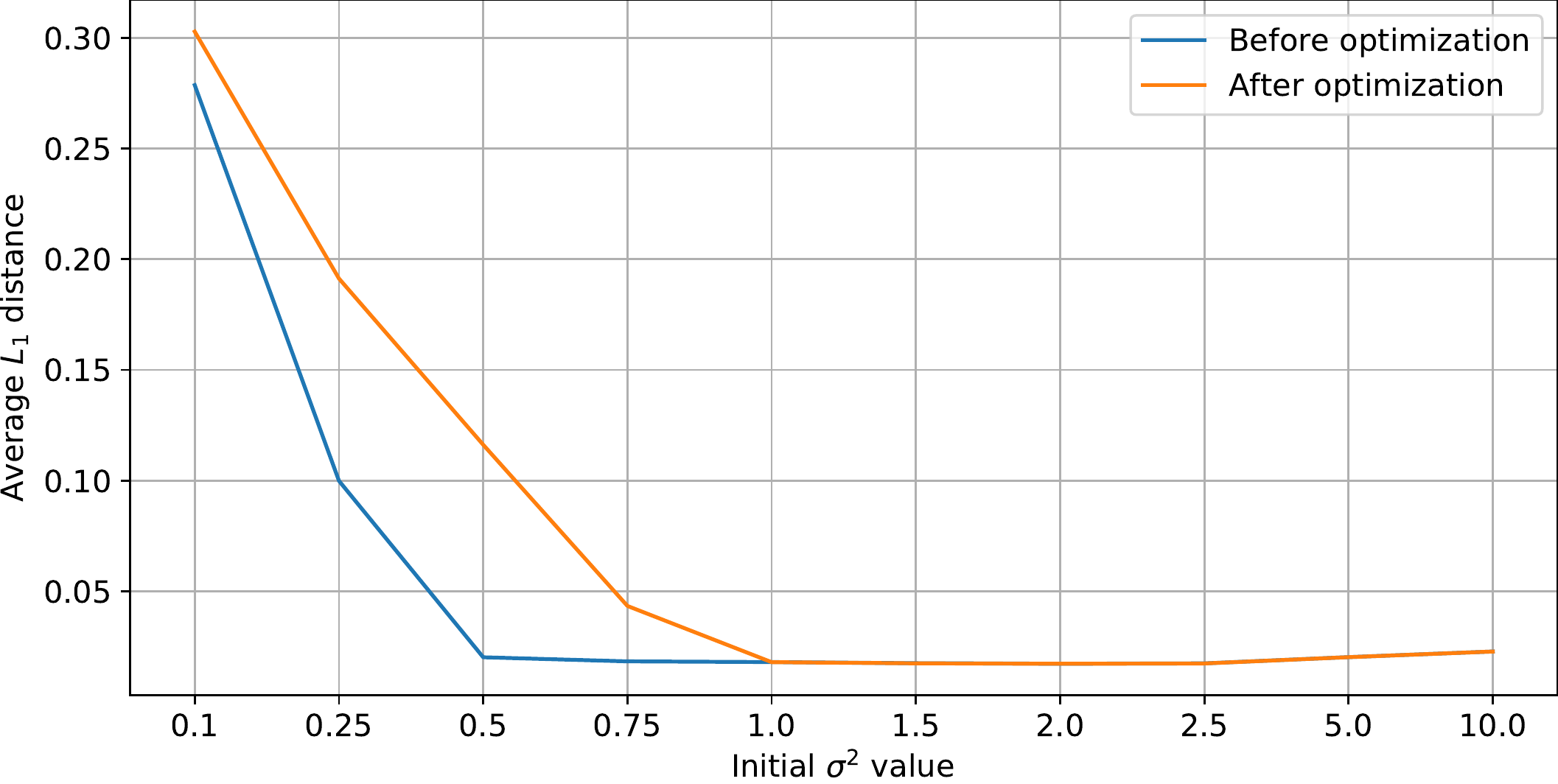}
    \caption{We tried to optimize the coordinates of the known points with gradient descent. As one can see, the optimization procedure diverges and it is much more important to properly pick the variance value $\sigma^2$ than to optimize the points.}
    \label{fig:optimizing-coords}
\end{figure}

\begin{figure}
    \centering
    \includegraphics[width=\linewidth]{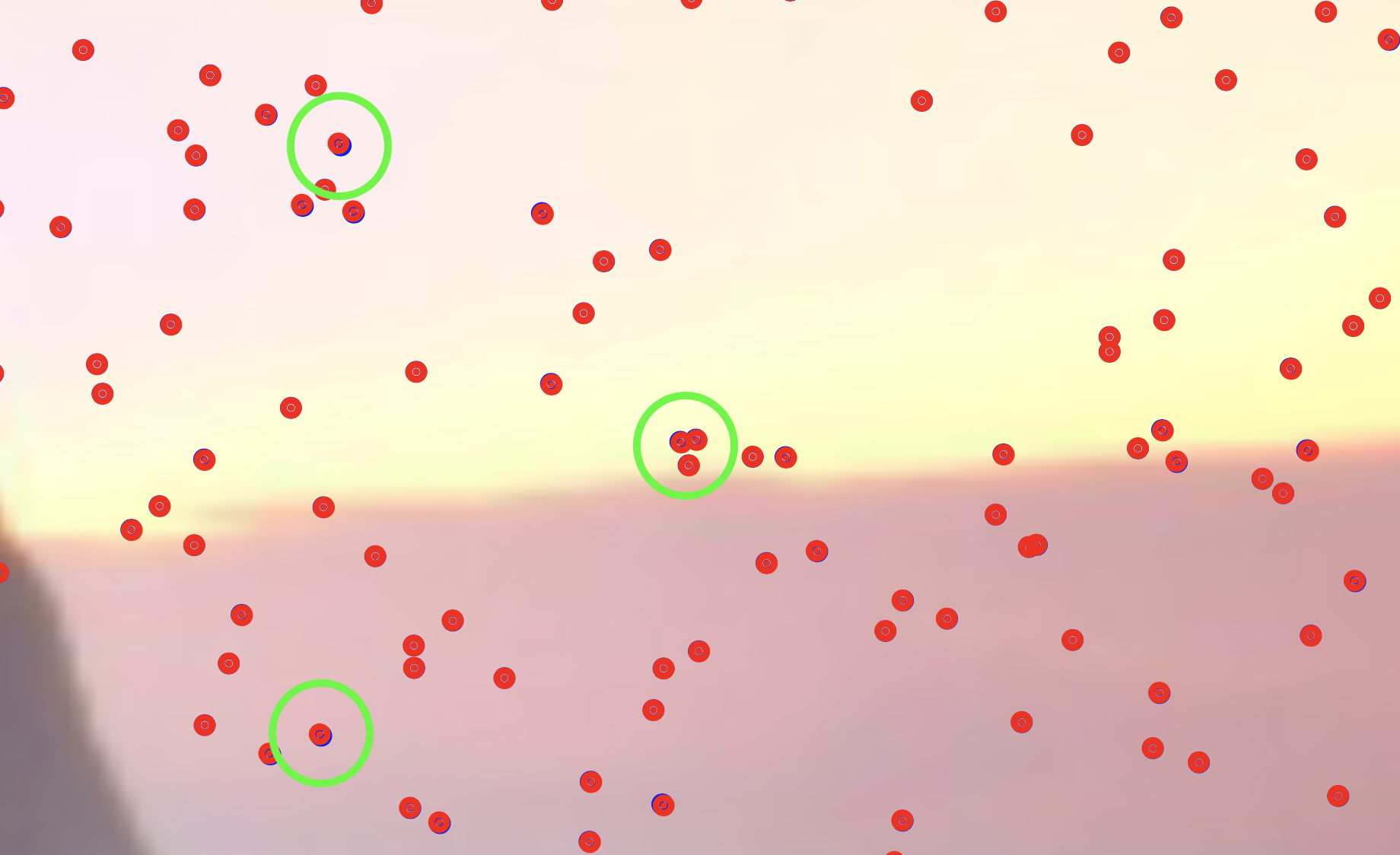}
    \caption{Known points positions before (blue) and after (red) the optimization procedure. The model is very reluctant to change the coordinates during the gradient descent $L_1$ minimization.}
    \label{fig:coords-paths}
\end{figure}

The results are presented on Figure~\ref{fig:optimizing-coords}.
As one can see, it is more important to select a proper variance value than optimizing the coordinates.
We hypothesize that the model is being stuck in a local minimum.
On Figure~\ref{fig:coords-paths}, we illustrate that the model is very reluctant to updating its coordinates positions.

\section{Conclusion}
In this work, we proposed an interpolation technique based on the gaussian mixture model which is able to reconstruct an image from arbitrary positioned points.
We developed an optimized CUDA kernel for both the forward procedure and the corresponding backward pass.
We benchmarked it against 6 existing interpolation techniques and showed that it outperforms them in terms of the reconstruction quality for a broad range of setups on ImageNet dataset.
Investigating why the model is not amenable to the optimization is a fruitful future research direction.

{\small
\bibliographystyle{ieee_fullname}
\bibliography{egbib}}

\newpage
\appendix

\end{document}